\documentclass{article}

\usepackage[final]{corl_2026} 

\usepackage{float}
\usepackage{wrapfig}
\usepackage{standalone}

\usepackage{graphicx}
\usepackage{booktabs}
\usepackage{multirow}
\usepackage{tabularx}
\usepackage{array}
\usepackage{amsmath}
\usepackage{amssymb}
\usepackage{xcolor}
\usepackage{enumitem} 
\usepackage[font=small,labelfont=bf]{caption}
\hypersetup{colorlinks=true, linkcolor=blue, citecolor=blue, urlcolor=blue}

\title{Catch Me If You Can:\\
Real-Time Feedback Denoising for Responsive VLAs}

\author{
  Yiheng Ji$^{1}$, Xingru Zhou$^{1}$, Luis Sentis$^{1*}$, Mingyo Seo$^{1  2*}$\\
  $^{1}$The University of Texas at Austin\\
  $^{2}$University of Central Florida
}

\begin{document}

\begingroup
\renewcommand{\thefootnote}{\fnsymbol{footnote}}
\maketitle
\footnotetext[1]{Equal advising and corresponding authorship.}
\endgroup

\begin{abstract}
Vision-Language-Action (VLA) models have shown strong generalization in robotic manipulation by combining semantic knowledge from pretrained vision-language models with expressive action-generation policies. Diffusion and flow-matching action generators are particularly effective for modeling temporally coherent action chunks, but these chunks are typically executed open-loop after inference. This limits responsiveness in dynamic manipulation when objects move, contacts change, or the scene evolves during execution.
We propose VLA-Feedback, a two-timescale architecture that combines
low-frequency diffusion planning with high-frequency visual feedback.
Rather than fully denoising an action chunk before execution,
VLA-Feedback retains its final denoising step as a lightweight feedback
interface, allowing each action to be corrected using the latest
observation before it is executed. This design preserves the expressiveness of the diffusion planner while enabling real-time action correction without rerunning the full vision-language diffusion model. VLA-Feedback largely preserved GR00T's static performance on LIBERO while improving average success on dynamic simulation tasks from \(27.5\%\) to
\(85.0\%\). On real-robot tasks, it improved average success from
\(53.3\%\) to \(73\%\). Additional materials can be found on our project page \url{https://vla-feedback.github.io}.
\end{abstract}


\begin{figure}[H]
    \centering
    \includegraphics[width=\textwidth]{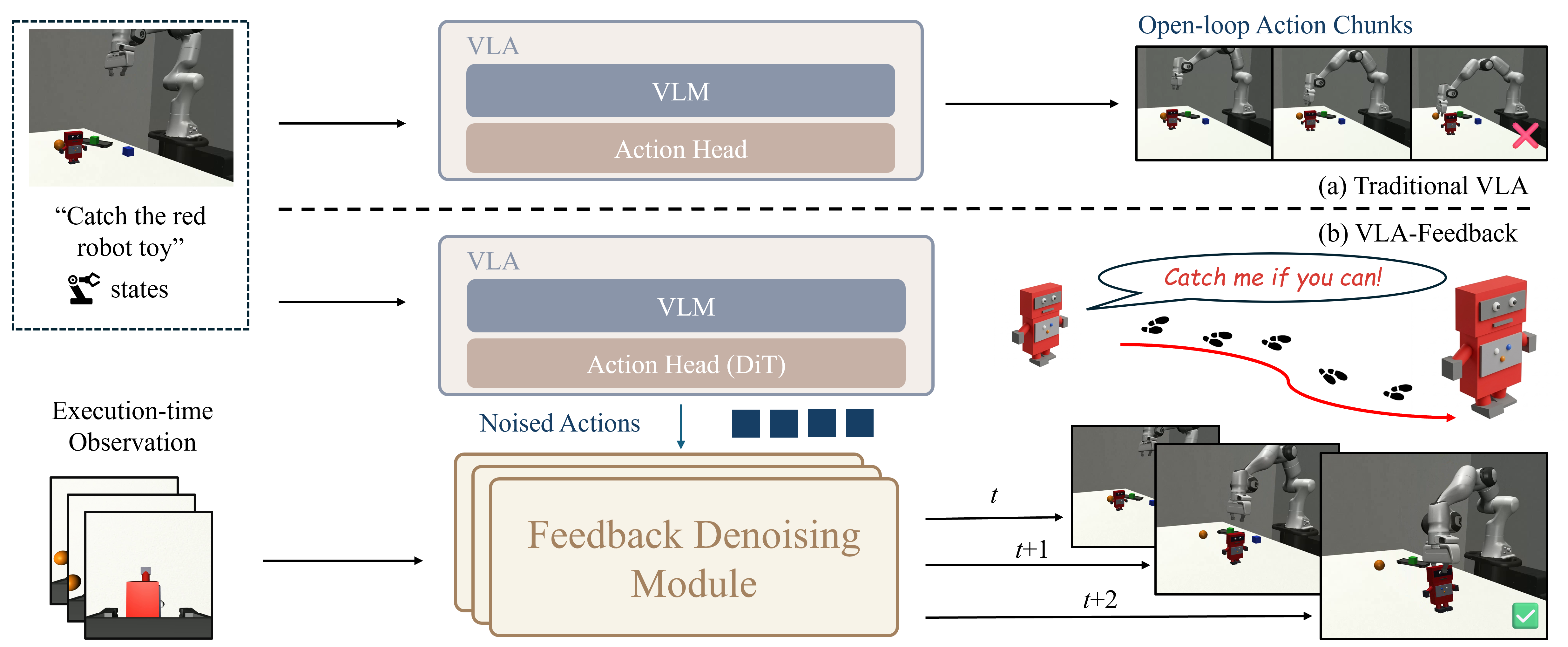}
    \captionsetup{font=footnotesize}
    \caption{\textbf{Overview of VLA-Feedback.}
    Traditional VLAs (a) execute a generated action chunk open-loop and
    therefore cannot incorporate observations received during execution.
    VLA-Feedback (b) retains the VLM-DiT planner in the slow pathway and uses
    a lightweight Feedback Denoising Module to correct each action using the
    latest observation, enabling intra-chunk feedback without rerunning the
    planner.}
    \label{fig:overview}
\end{figure}

\section{Introduction}

Vision-Language-Action (VLA) models have emerged as a promising paradigm for language-conditioned robotic manipulation~\cite{pmlr-v229-zitkovich23a,pmlr-v270-kim25c}. By combining pretrained vision-language representations with robot action generation, these models can leverage semantic knowledge while producing low-level control commands for diverse manipulation tasks. Even with strong semantic understanding, VLAs can struggle with dynamic
tasks because their computational cost prevents full-model inference at
control frequency ~\cite{jiang2026fast,agouzoul2026understandingasynchronousinferencemethods}. Action chunking~\cite{black2026pi0visionlanguageactionflowmodel,11247519} is an effective way to reduce end-to-end execution latency by producing a sequence of future actions with a single inference call, allowing inference and action execution to be amortized or overlapped~\cite{jiang2026fast}. However, it still operates in an open-loop manner, where new sensory observations during action chunk execution are ignored~\cite{liang2026adaptiveactionchunkinginferencetime,lu2026fasterrethinkingrealtimeflow}. This becomes problematic when the robot must react quickly to changes in the environment, for example when picking up a can whose rolling motion is caused either by environmental factors, such as an uneven table, or by perturbations from the robot itself.

Inspired by hierarchical or two-timescale approaches~\cite{chen2025fast,han2024dual,bu2024towards,tang2025vlashrealtimevlasfuturestateaware}, one line of work to address latency is to separate slow high-level reasoning from fast low-level action execution, where a large slow vision-language model (VLM) and a lightweight fast action module run at different frequencies. The decoupling of high-level reasoning and low-level control enables high inference efficiency and rapid reaction to new observations. However, these methods face a latency-capacity trade-off: If the action head is too lightweight, such as a shallow multilayer perceptron (MLP), it may sacrifice the expressive action generation capability; if the fast pathway is made more expressive, for example with a diffusion-based action module, its latency increases and action chunking is again needed for efficiency~\cite{agouzoul2026understandingasynchronousinferencemethods}.

Another approach is to improve action generation by conditioning it on richer observations or task constraints during generation. In diffusion models, this can be implemented through guidance mechanisms or conditional denoising branches, represented by classifier-free guidance and ControlNet-style architectures~\cite{ho2022classifier,Zhang_2023_ICCV}. These methods show that intermediate denoising states can be steered by additional information, suggesting a natural strategy for robotic action generation: use richer conditions to produce better action chunks. However, such conditioning is applied only when the chunk is initially
generated; it cannot incorporate observations received later during
chunk execution. Thus, for action-chunking VLAs, the key challenge is not only how to generate a better action chunk, but how to keep an already generated chunk responsive during execution~\cite{liang2026adaptiveactionchunkinginferencetime,lu2026fasterrethinkingrealtimeflow}.

Our key insight is to perform feedback correction within the generative
action process. Rather than generating an independent action with a
separate reactive policy or applying a post-hoc residual to the final
planner output, we condition the planner's final denoising step on the
latest observation. This keeps the corrected action anchored to the
planner's learned action structure while enabling intra-chunk feedback (Fig.~\ref{fig:overview}).

To this end, we propose VLA-Feedback, a two-timescale architecture for responsive diffusion-based VLAs. It keeps the VLM-DiT planner in the slow pathway, where it produces a near-final action chunk at sparse planning intervals. During execution, a fast feedback module reuses this structured intermediate representation and predicts the final denoising velocity conditioned on the latest observation. By exposing the final denoising transition as a lightweight feedback interface, VLA-Feedback corrects actions at control frequency without rerunning the full vision-language diffusion planner. This directly addresses the open-loop limitation of action chunking while retaining the capacity of diffusion-based action generation. In simulation, our method largely preserved GR00T's static performance on LIBERO while improving average dynamic-task success from 27.5\% to 85.0\%. On real hardware, it improved average success from 53.3\% to 73\% across static and dynamic manipulation tasks.





\section{Related Work}

\paragraph{Efficient VLA Inference.}
Prior work has made substantial progress in improving VLA efficiency~\cite{guan2025efficientvisionlanguageactionmodelsembodied,yu2026surveyefficientvisionlanguageactionmodels} through smaller vision-language backbones~\cite{shukor2025smolvla,10900471,budzianowski2025edgevlaefficientvisionlanguageactionmodels}, lightweight action heads~\cite{dong2025vitavlaefficientlyteachingvisionlanguage}, token reduction~\cite{tan2025thinktwiceactonce,PertschK-RSS-25}, model compression~\cite{NEURIPS2025_3a2ef31a,wang2026specprunevla,pei2026action}, and action chunking~\cite{11247519}. Asynchronous ``thinking while acting'' designs are especially effective at reducing blocking latency by allowing slow reasoning and fast action execution to proceed in parallel~\cite{NEURIPS2025_300ccb21,xie2026dynamicvlavisionlanguageactionmodeldynamic,jiang2026asyncvlaasynchronousflowmatching,wang2026discretertcdiscretediffusionpolicies}. However, these methods either trade off model capacity, execute generated chunks open-loop, or require additional mechanisms to handle stale plans and inconsistencies across planning boundaries.

\paragraph{Diffusion-Based Action Generation.}
 Diffusion models were introduced to robotics for modeling multimodal action distributions and generating smooth trajectories~\cite{pmlr-v162-janner22a,chi2025diffusion,10342382,pmlr-v270-yang25a}. Faster samplers and training objectives, such as DDIM, distillation, consistency models, and flow matching, reduce the cost of iterative generation~\cite{song2021denoising,salimans2022progressive,pmlr-v202-song23a,lipman2023flow,Prasad-RSS-24,pmlr-v267-wang25ba}. Recent VLA models further use diffusion or flow-matching action heads to generate horizon-length action chunks from vision-language representations~\cite{black2026pi0visionlanguageactionflowmodel,ICLR2025_49f80e4d,nvidia2025gr00tn1openfoundation,pmlr-v305-wen25b,Hou_2025_ICCV}. Diffusion guidance, including classifier-free guidance and ControlNet-style conditioning, has also been widely studied in computer vision~\cite{ho2022classifier,Zhang_2023_ICCV} and explored in robotics~\cite{pmlr-v305-li25c,11272538,10912754,11127231}. Nevertheless, these methods either still incur non-negligible sampling latency or leave the generated action chunk open-loop during execution.

\paragraph{Dual-System VLAs and Feedback Correction.}
Motivated by \cite{kahneman2011thinking}, dual-system VLAs usually consist of a System 2 module, often a VLM, for high-level planning, and a System 1 module for generating actions conditioned on the VLM outputs~\cite{Brohan-RSS-23,pmlr-v229-zitkovich23a,pmlr-v270-kim25c,Ghosh-RSS-24,song2025hume}. These methods are not primarily optimized for low-latency control, since the fast action module still depends on the slow reasoning module and must wait for its outputs. To improve responsiveness, recent works run the two systems at different frequencies, with System 1 operating at a higher control rate~\cite{pmlr-v270-zhang25b,han2024dual,chen2025fast,bu2024towards,xiong2026hypervla}. However, these methods still face a latency-capacity tradeoff or leave the generated action chunk open-loop during execution. A related strategy is to use a high-frequency decorator to correct policy outputs, such as residual action correction methods~\cite{sendai2025leave,yuan2025policy}, but these corrections are typically applied in final action space, and may not preserve the action prior learned by the planner. In contrast, VLA-Feedback does not generate a separate fast action or apply a post-hoc residual; it performs observation-conditioned denoising velocity updates on the slow planner's near-final action.

\begin{figure}[!t]
    \centering
    \includegraphics[width=\textwidth]{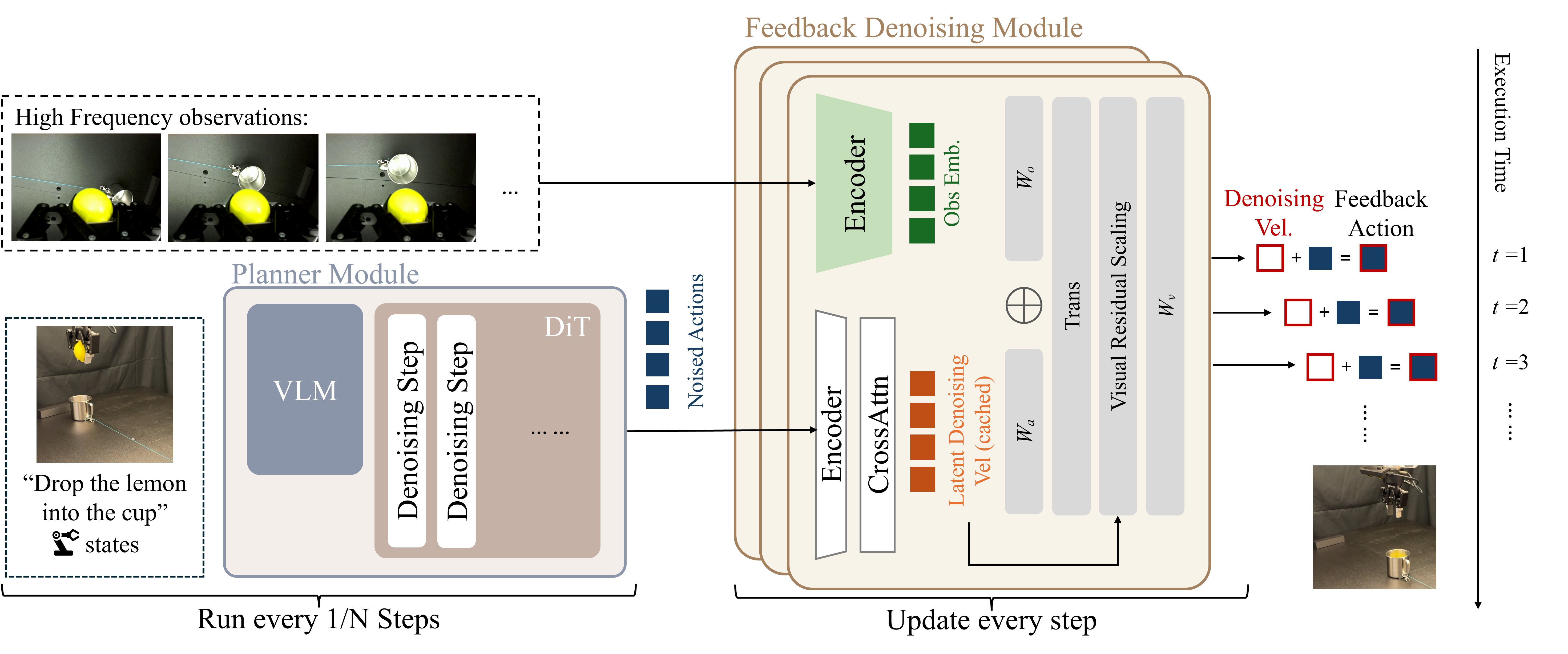}
    \captionsetup{font=footnotesize}
    \caption{\textbf{Architecture of VLA-Feedback.} A low-frequency VLM-DiT planner generates a near-final action chunk and caches its encoded action representation. At every control step, the Feedback Denoising Module fuses the cached action representation with the latest high-frequency observation, applies visual residual scaling, and predicts the final denoising velocity to refine the current action online.}
    \label{fig:architecture}
\end{figure}

\section{Method}
\subsection{Problem Formulation}

We model language-conditioned manipulation as a partially observed Markov decision process $\mathcal{M}=(\mathcal{S},\mathcal{O},\mathcal{A},P,R,\gamma)$. Here, $\mathcal{S}$, $\mathcal{O}$, and $\mathcal{A}$ denote
the state, observation, and action spaces, respectively;
$P$ is the state-transition model, $R$ the reward function,
and $\gamma$ the discount factor. At each timestep $t$, the robot receives an observation $o_t \in \mathcal{O}$ of the underlying state $s_t \in \mathcal{S}$ and an observed robot state $q_t$. A policy $\pi$ maps the observation, robot state, and language instruction $\ell$ to an action $a_t \in \mathcal{A}$. The objective is to maximize the expected discounted return $\mathbb{E}_{\pi}\left[\sum_{t=0}^{\infty}\gamma^t R(s_t,a_t,s_{t+1})\right]$. For efficient VLA inference, action-chunking policies predict an action sequence where $H$ denotes the action-chunk horizon and $t$ denotes the timestep:
\begin{equation}
\mathbf{a}_{t:t+H-1} = \pi(o_t, q_t, \ell).
\label{eq:action_chunk}
\end{equation}
The predicted chunk is then executed over the next \(H\) control steps. However, standard action chunking is open-loop within the chunk: although the robot receives new observations \(o_{t+h}\) during execution, these observations are not used to update the already generated actions. We formulate feedback-augmented action chunking to address this limitation. At planning step \(t\), a slow planner generates a near-final action-space trajectory \(\mathbf{a}^{k-1}_{t:t+H-1}\). During execution, a feedback module refines the corresponding action using the latest observation,
\begin{equation}
a^k_{t+h}=g(a^{k-1}_{t+h}, o_{t+h}), \quad h\in\{0,\ldots,H-1\}.
\label{eq:feedback_refinement}
\end{equation}
Here, \(g(\cdot)\) denotes the complete Feedback Denoising Module, which performs an observation-conditioned final denoising update rather than generating a separate action from scratch. This allows expensive VLA planning to run at low frequency while keeping the action chunk responsive to execution-time observations.

\subsection{VLA-Feedback Architecture}
Figure~\ref{fig:architecture} illustrates VLA-Feedback, which combines a low-frequency diffusion planner with a high-frequency feedback denoising module. We first formalize feedback-augmented action chunking, then describe the architecture and training/inference procedure.

\paragraph{Planner Module.}
We instantiate the high-level planner using the GR00T foundation model, which integrates a VLM with a diffusion-based action generator and provides a strong pretrained VLA backbone~\cite{nvidia2025gr00tn1openfoundation}. We additionally evaluate an implementation with a $\pi_0$~\cite{black2026pi0visionlanguageactionflowmodel} planner in Section~\ref{sec:ablation}. Given a low-frequency observation \(o_t\), a language instruction \(\ell\), and the current observed robot state \(q_t\), a pretrained NVIDIA Eagle-2 model encodes the visual-language input into a latent task representation \(z_t\)~\cite{li2025eagle2buildingposttraining}. Conditioned on \(z_t\) and \(q_t\), a Diffusion Transformer (DiT) performs iterative denoising to generate an action trajectory over a fixed horizon \(H\)~\cite{Peebles_2023_ICCV}.

Let \(k\) index the final denoising step, with \(\mathbf{a}^{k-1}\) denoting the near-final action trajectory and \(\mathbf{a}^{k}\) denoting the final action trajectory. In a standard diffusion-based VLA, the planner completes all denoising steps and directly outputs \(\mathbf{a}^{k}_{t:t+H-1}\) for execution. In contrast, VLA-Feedback stops before the final denoising step and outputs a near-final action-space trajectory \(\mathbf{a}^{k-1}_{t:t+H-1}\), which is passed to the fast Feedback Denoising Module. This design keeps the expressive diffusion planner while exposing the last denoising step for execution-time observation feedback.

\paragraph{Feedback Denoising Module.}
At planning step \(t\), after the DiT produces the near-final action-space trajectory \(\mathbf{a}^{k-1}_{t:t+H-1}\), the Feedback Denoising Module first computes an action-side representation for the whole chunk. This action-side computation is performed once together with the slow planner and cached during execution. At each execution step \(t+h\), where \(h \in \{0,\ldots,H-1\}\), the module uses the latest observation \(o_{t+h}\) to perform the final denoising update for the corresponding near-final action \(a^{k-1}_{t+h}\).

The module contains two branches. The action branch encodes the near-final action-space trajectory using an action encoder, cross-attention module, and a single-layer projection, while the observation branch encodes the real-time observation using a lightweight visual encoder \(\phi(\cdot)\) followed by another single-layer projection~\cite{NIPS2017_3f5ee243}:
    \begin{equation}
    \begin{aligned}
    \mathbf{e}^a_{t:t+H-1} &= W_a\!\left(\mathrm{CrossAttn}_a\!\left(\mathrm{Enc}_a(\mathbf{a}^{k-1}_{t:t+H-1})\right)\right), \\
    e^o_{t+h} &= W_o\!\left(\phi(o_{t+h})\right).
    \end{aligned}
    \label{eq:feedback_features}
    \end{equation}
Here, \(\mathbf{e}^a_{t:t+H-1}\) is computed once per action chunk and cached, while \(e^o_{t+h}\) is updated at every control step. At execution step \(t+h\), the cached action feature \(e^a_{t+h}\) and the current observation feature \(e^o_{t+h}\) are fused using a lightweight transformer/cross-attention block~\cite{NIPS2017_3f5ee243}:
    \begin{equation}
    h_{t+h}=\mathrm{Trans}(e^a_{t+h}, e^o_{t+h}).
    \label{eq:feature_fusion}
    \end{equation}
To control how strongly visual feedback changes the planner output, we use a visual residual scale that interpolates between the original action embedding and the observation-conditioned transformer output:
    \begin{equation}
    h^{fb}_{t+h}=e^a_{t+h}+\alpha(h_{t+h}-e^a_{t+h}),
    \label{eq:visual_residual}
    \end{equation}
    where \(\alpha\) is a learnable scale. This residual scaling anchors the feedback representation to the near-final action proposed by the diffusion planner while allowing the latest observation to adjust the final denoising direction. Finally, a single-layer velocity head predicts the observation-conditioned feedback denoising velocity \(\hat{v}^{fb}_{t+h}=W_v(h^{fb}_{t+h})\), and the final action command is obtained through one denoising update with step size \(\Delta k\):
    \begin{equation}
    a^k_{t+h}=a^{k-1}_{t+h}+\Delta k \hat{v}^{fb}_{t+h}.
    \label{eq:feedback_denoising_update}
    \end{equation}
Therefore, VLA-Feedback does not train a separate high-frequency policy or apply a post-hoc residual after action generation. Instead, it reuses the slow planner's near-final action trajectory and performs observation-conditioned denoising velocity updates during execution.

\subsection{Training and Inference Strategy}
We adopt a two-phase training strategy to decouple long-horizon diffusion planning from execution-time feedback refinement. The planner and feedback module are trained sequentially to isolate their respective roles. In the first phase, we train the planner using the standard flow-matching objective adopted in GR00T~\cite{nvidia2025gr00tn1openfoundation,lipman2023flow}. The Diffusion Transformer predicts a velocity field \(\hat{v}\), which is supervised by the target velocity \(v\) derived from expert actions. The planner is optimized using a masked regression loss:
\begin{equation}
\mathcal{L}_1
=
\|\hat{v} - v\|_2^2 .
\label{eq:planner_loss}
\end{equation}

In the second phase, we freeze the pretrained planner, including the VLM, DiT, and the shared action-side denoising branch used by the Feedback Denoising Module. We then train only the newly introduced feedback components. For each training sample, we first run the planner to obtain the near-final action-space trajectory \(\mathbf{a}^{k-1}_{t:t+H-1}\) and the cached action-side features \(\mathbf{e}^{a}_{t:t+H-1}\). The feedback module then uses the cached action feature \(e^a_{t+h}\) and the high-frequency observation \(o_{t+h}\) to predict the feedback denoising velocity \(\hat{v}^{fb}_{t+h}\) and produce the refined final action \(\hat{a}^{k}_{t+h}\). The feedback module is trained with supervised regression to the ground-truth action \(a_{t+h}^{\mathrm{gt}}\):
\begin{equation}
\mathcal{L}_2
=
\mathbb{E}\left[
\left\|\hat{a}^{k}_{t+h} - a_{t+h}^{\mathrm{gt}}\right\|_2^2
\right].
\label{eq:feedback_loss}
\end{equation}
This phase teaches the feedback module to correct the planner's near-final actions using real-time observations, without modifying the planner.


At test time, the planner runs once per chunk to produce a horizon-\(H\)
near-final action trajectory and cached action-side features. Before each
of the \(H\) actions is executed, the feedback module uses the latest
observation to perform the final denoising update, without re-invoking
the planner. Thus, each chunk follows a \(1{:}H{:}H\)
planner–feedback–action schedule (\(1{:}16{:}16\) in our experiments).

\section{Experiments}
We organized the experiments around four research questions:
(i) Could VLA-Feedback preserve the static manipulation capability of the diffusion planner?
(ii) Could it improve robustness when the scene changed during action-chunk execution?
(iii) Could high-frequency feedback reduce the delay before new observations affected actions?
(iv) How important were feedback frequency and denoising-space refinement compared with two-timescale dual-system updates and direct action residual correction?
We studied these questions using static LIBERO tasks, dynamic simulation tasks, real-world hardware experiments, reaction-latency analysis, and ablations.

\subsection{Experiment Setup}

\begin{figure}[!t]
\includegraphics[width=\textwidth]{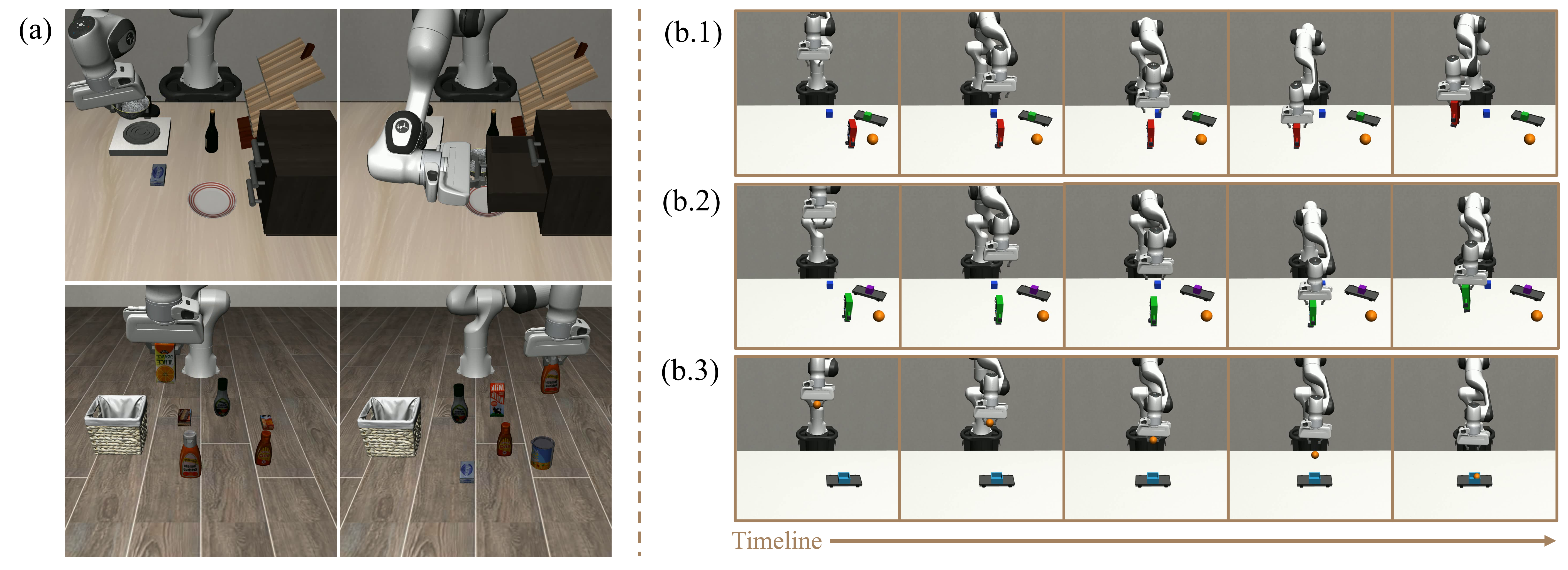}
\caption{\textbf{Simulation tasks.} (\textbf{a}) Example tasks from the LIBERO benchmark: the top two were from LIBERO-Goal, and the bottom two were from LIBERO-Object. (\textbf{b}) Time-lapse rollouts of VLA-Feedback on three dynamic tasks: ``Pick up the red robot toy,'' ``Pick up the green robot toy,'' and ``Drop the ball into the cup.''}
\label{fig:sim}
\end{figure}

\paragraph{Simulation Setup.}
Static simulation used LIBERO-Object and LIBERO-Goal~\cite{liu2023libero} to test whether feedback refinement preserved standard language-conditioned manipulation. Dynamic simulation used three Robosuite~\cite{zhu2020robosuite} tasks: grasping a toy robot whose speed and direction changed randomly every 2 seconds during execution, and dropping a ball into a moving cup (Fig.~\ref{fig:sim}). These tasks intentionally created mid-chunk target displacement, so an action chunk planned at the beginning could become outdated before execution finished. We further evaluated four unseen dynamic variants that changed object appearance, object shape, motion pattern, or speed range.

\paragraph{Real-world Setup.}
We used a Franka Emika Panda robot on one static task and two dynamic tasks. The static task evaluated normal grasping, while the dynamic tasks involved catching a rolling can and dropping a lemonade into a moving cup. These tasks tested whether feedback could maintain spatial alignment and timing under real-world motion uncertainty.

\paragraph{Training and Evaluation.}
All real-world policies were trained with 50 demonstrations per task and evaluated over 20 rollouts. In simulation, policies were trained with 50 demonstrations per LIBERO task and 60 demonstrations per dynamic task, and were evaluated over 20 rollouts per LIBERO task and 40 rollouts per dynamic task. Detailed task descriptions, success criteria, and motion-generation procedures are provided in Appendix~\ref{app:tasks}. For fair comparison, all methods used the same demonstrations, evaluation splits, initial-state distributions, target-motion patterns, rollout budgets, and success criteria. We compared against OpenVLA, GR00T, and FiS-VLA using their released finetuning protocols. GR00T and VLA-Feedback shared the same VLM-DiT planner, action horizon, and control frequency; VLA-Feedback differed only by replacing the final denoising step with the feedback denoising module. Full baseline details appear in Appendix~\ref{app:baselines}. We used GR00T as the default backbone to isolate the effect of feedback denoising through matched comparisons with its original planner; the $\pi_0$-based implementation was evaluated separately to test applicability across backbones (Table~\ref{tab:pi0}).


\subsection{Performance on Simulation Tasks} \label{sec:sim}

\begin{table}[t]
\centering
\caption{Success rates on static LIBERO, dynamic tasks, and unseen tasks}
\label{tab:sim_results}
\makeatletter\def\@captype{table}
\resizebox{\columnwidth}{!}
{
\begin{tabular}{lcccccccccccc}
\toprule
Method & LIBERO & LIBERO& Pick up 1D & Pick up 2D & Drop Ball to & Catch & 1D Robot  & Drop Ball & Drop Ball  \\
& Goal & Obj. & Robot & Robot & Cup & Block$^\ast$ & Speed$^\ast$ & Color$^\ast$  &  Speed$^\ast$ \\
\midrule
OpenVLA~\cite{pmlr-v270-kim25c} & 78 & 88.5 & 5 & 60 & 0 & 0 & 0 & 0 & 0 \\
FiS-VLA~\cite{chen2025fast} & 41.5 & 53.5 & 0 & 67.5 & 80 & 0 & 0 & 60 & 0 \\
GR00T~\cite{nvidia2025gr00tn1openfoundation} & \textbf{92} & \textbf{97.5} & 47.5 & 20 & 15 & 60 & 0 & 2.5 & 0 \\
VLA-Feedback & \textbf{92} & 95.5 & \textbf{80} & \textbf{75} & \textbf{100} & \textbf{67.5} & \textbf{50} & \textbf{97.5} & \textbf{57.5} \\
\bottomrule
\end{tabular}
}
{\raggedright\footnotesize $^\ast$ unseen dynamic tasks.\par}
\end{table}

\paragraph{Static and Dynamic Tasks.}
Table~\ref{tab:sim_results} showed that VLA-Feedback largely preserved the static-task performance of the diffusion planner: it matched GR00T on Goal and was 2 percentage points lower on Object. This suggested that the feedback decoder maintained comparable performance on mostly stationary scenes.

The main gain appeared on dynamic tasks, where the target could move after the initial action chunk was generated. In this setting, chunk-based policies could execute stale actions based on outdated observations. VLA-Feedback kept the same VLM-DiT planner but updated the final denoising velocity using the latest observation at action time. The larger gap on dynamic tasks compared with static LIBERO tasks suggested that the improvement came from intra-chunk feedback.



\paragraph{Unseen Tasks: Generalization Experiments.}
VLA-Feedback performed best across all unseen dynamic variants in Table~\ref{tab:sim_results}. Since these variants changed the object, appearance, motion pattern, or speed distribution, the improved performance suggested that VLA-Feedback was not only memorizing a single fixed motion pattern, but could use online visual feedback to correct the near-final action under execution-time uncertainties when the target behavior differed from the demonstrations.

\begin{wraptable}{r}{0.45\textwidth}
\centering
\vspace{3pt}
\caption{\textbf{Timing and reaction latency of VLA policies.} S:F:A denotes slow-planner calls, fast-module updates, and executed actions per chunk.}
\label{tab:reaction_latency}
\scriptsize
\setlength{\tabcolsep}{3pt}
\renewcommand{\arraystretch}{1.05}
\begin{tabular}{@{}lccccc@{}}
\toprule
Method & S:F:A & $T_s$ & $T_f$ & $R$ & $\bar{T}_{\mathrm{react}}$ \\
\midrule
OpenVLA~\cite{pmlr-v270-kim25c} & $1{:}0{:}1$ & 160 ms & -- & 1 & $160+\frac{1}{2}\Delta t$ \\
GR00T~\cite{nvidia2025gr00tn1openfoundation} & $1{:}0{:}16$ & 80 ms & -- & 16 & $80+8\Delta t$ \\
FiS-VLA~\cite{chen2025fast} & $1{:}4{:}4$ & 73 ms & 40 ms & 1 & $40+\frac{1}{2}\Delta t$ \\
VLA-Feedback & $1{:}16{:}16$ & 79 ms & 2 ms & 1 & $2+\frac{1}{2}\Delta t$ \\
\bottomrule
\end{tabular}
\end{wraptable}

\paragraph{Reaction Latency Analysis.}
We estimated reaction latency as the delay between a new observation becoming available and the first executed action that could depend on it.
This metric captured responsiveness rather than throughput.
Let $R$ be the number of executed actions between two observation-conditioned updates, $T_{\mathrm{update}}$ be the runtime of the module that incorporates the new observation, and $\Delta t$ be the control interval.
Assuming changes arrive uniformly between update opportunities, the expected waiting time is $\frac{R}{2}\Delta t$, giving
\begin{equation}
    \bar{T}_{\mathrm{react}}
    =
    T_{\mathrm{update}} + \frac{R}{2}\Delta t .
    \label{eq:reaction_latency}
\end{equation}

For single-system policies, $T_{\mathrm{update}}=T_{\mathrm{slow}}$.
For two-system policies, $T_{\mathrm{update}}=T_{\mathrm{fast}}$, while $R$ depends on how often the fast pathway updates the executed action.
The S:F:A column in Table~\ref{tab:reaction_latency} reported the number of slow-planner calls, fast-module updates, and executed actions within one action chunk; for example, $1{:}16{:}16$ meant one slow planner call, 16 feedback updates, and 16 executed actions. VLA-Feedback achieved the lowest expected reaction latency because its 2 ms feedback denoising module updated every action step.

\subsection{Performance on Real-World Tasks} \label{sec:real}

\begin{figure}[t]
    \centering
    \includegraphics[width=\textwidth]{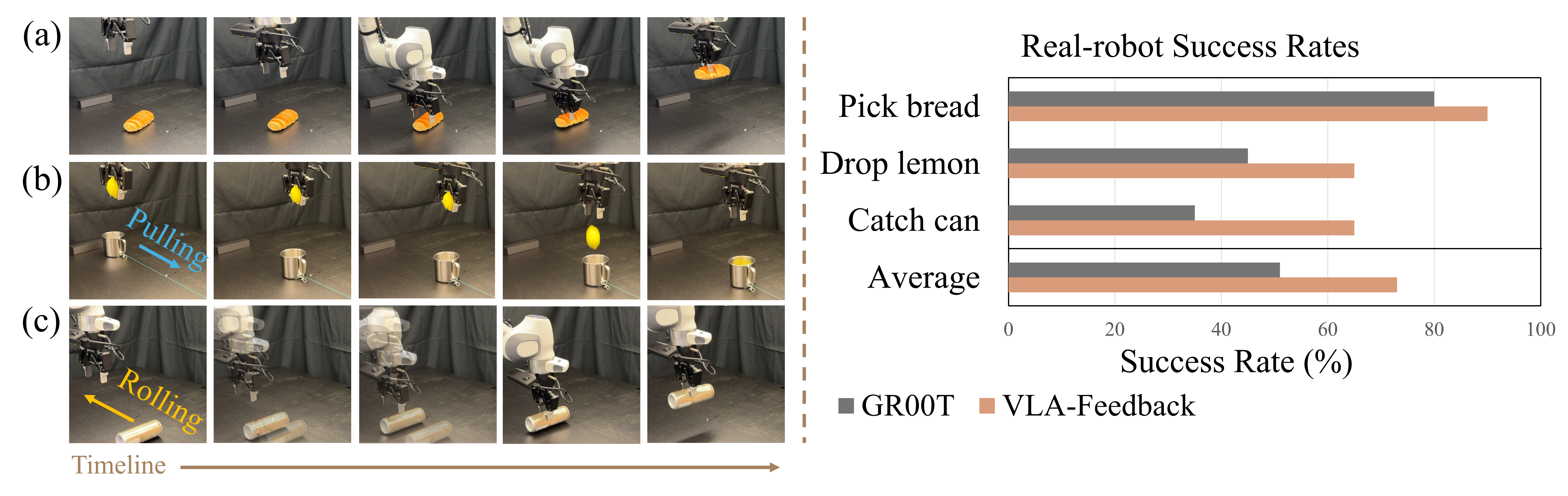}
    \caption{\textbf{Real-robot experiments on Panda.} (\textbf{Left}) execution timelines for three tasks: (a) ``Pick up the bread,'' (b) ``Drop the lemonade into the cup,'' and (c) ``Catch the rolling can.'' (\textbf{Right}) success rates across the three tasks.}
    \label{fig:realrobot}
\end{figure}

Figure~\ref{fig:realrobot} showed that the simulation trend also held on hardware. On the static bread-picking task, VLA-Feedback slightly improved over GR00T, suggesting that feedback refinement did not destabilize manipulation when the target remained stationary. The gains were larger on the two dynamic tasks, where the target changed during execution and the original action chunk could become stale. This gap suggested that GR00T often acted on outdated target observations, while VLA-Feedback could use the latest hand-view observation to update grasp alignment, release position, and timing. The real-world gains were smaller than in simulation, which was expected because object motion was less repeatable and the observation was affected by perception noise or actuation delay.


\subsection{Ablation and Additional Studies} \label{sec:ablation}


\begin{figure*}[!t]
    \centering
    \begin{minipage}[t]{0.495\textwidth}
        \centering
        \includegraphics[width=\textwidth]{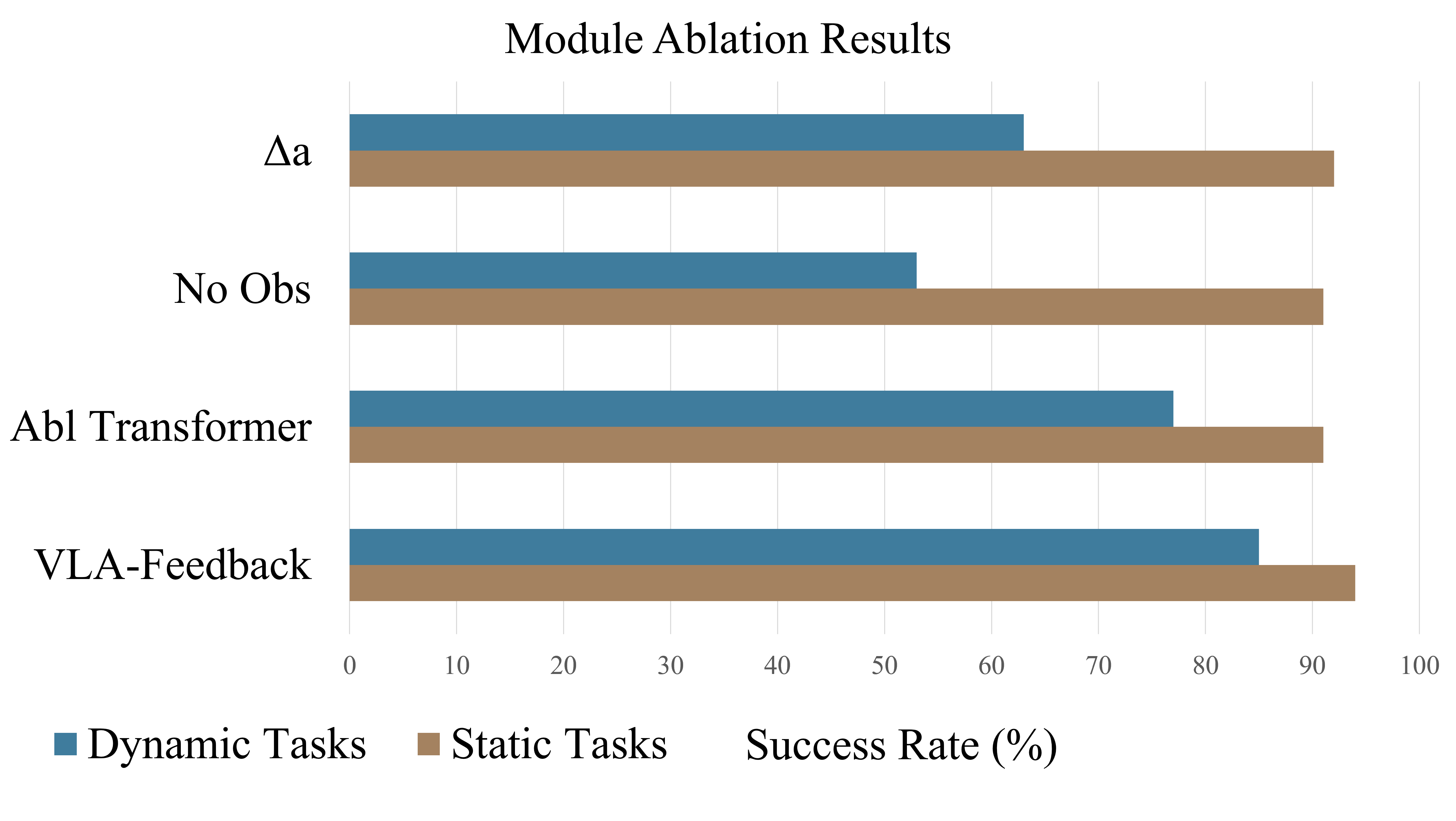}
    \end{minipage}
    \hfill
    \begin{minipage}[t]{0.495\textwidth}
        \centering
        \includegraphics[width=\textwidth]{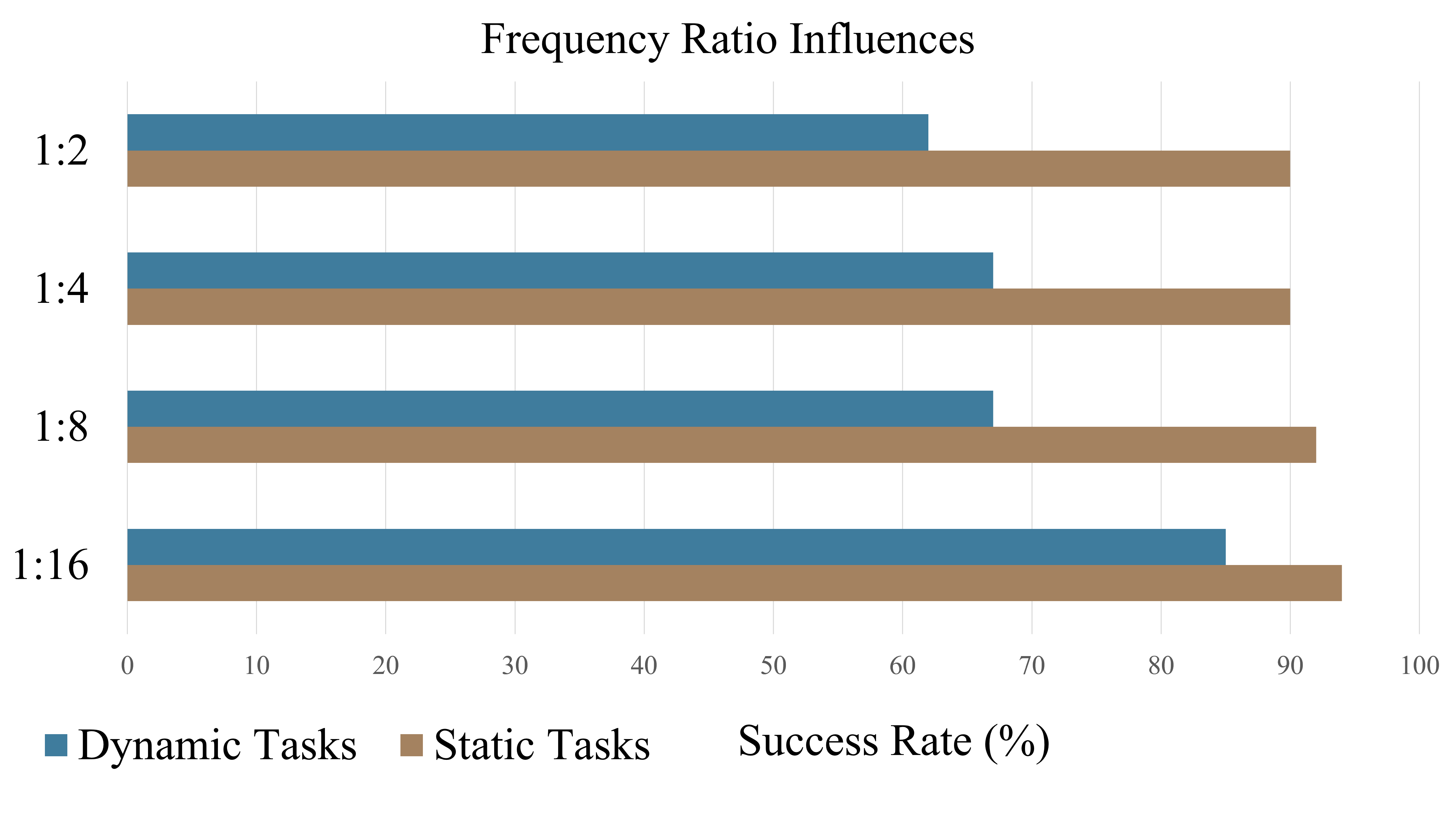}
    \end{minipage}
    \captionsetup{font=footnotesize}
    \caption{\textbf{Ablation study of VLA-Feedback.} (\textbf{a}) Ablations compared VLA-Feedback with direct action residual correction, no visual feedback, and replacement of the Transformer fusion block with an MLP. Removing visual feedback or using direct action residual correction most strongly degraded dynamic-task performance, while static performance remained similar. (\textbf{b}) Reducing the feedback update frequency lowered dynamic-task success rate, showing that frequent intra-chunk feedback was important for non-stationary manipulation.}
    \label{fig:ablation}
\end{figure*}

\paragraph{Ablation on Modules.}
Figure~\ref{fig:ablation} (a) tested three design choices: whether structured visual-action fusion was needed, whether the gain came from real-time observation input, and whether denoising-space refinement was better than direct action residual correction. Replacing the transformer fusion block with an MLP slightly reduced performance, suggesting that structured fusion between the near-final action and visual feedback was useful. Removing the observation input caused a much larger drop on dynamic tasks, showing that the improvement depended on real-time visual feedback. Directly predicting an action-space residual $\Delta a$ performed worse than denoising-velocity refinement, especially on dynamic tasks. One possible reason was that the planner output was already near-final, so direct residual targets were small and could be noisy to learn. In contrast, denoising-velocity refinement gave the feedback module a more structured correction interface, allowing visual feedback to adjust the planner's final denoising direction rather than post-process the final action. These results directly addressed RQ4 by validating the proposed feedback and refinement design.

\paragraph{Ablation on Two-System Frequency.}
Figure~\ref{fig:ablation} (b) tested whether high-frequency feedback was necessary or whether slower feedback was sufficient. The default ratio of 1:16 ran the VLM-DiT planner once and applied 16 feedback updates during execution. Reducing the feedback frequency had a small effect on static tasks but sharply reduced dynamic-task success. This supported the design choice: VLA-Feedback was most useful when the scene changed after the initial diffusion plan was generated, and weakening feedback mainly hurt those settings. Thus, the dynamic gains came not only from adding a feedback decoder, but from applying feedback at action-level frequency.

\begin{wraptable}{r}{0.45\textwidth}
\centering
\caption{\textbf{Evaluation across VLA foundation model backbones.}
Success rates (\%) on three dynamic tasks.}
\label{tab:pi0}
\scriptsize
\setlength{\tabcolsep}{3pt}
\renewcommand{\arraystretch}{1.05}
\begin{tabular}{@{}lccc@{}}
\toprule
Method & 1D Robot & 2D Robot & Drop Ball \\
\midrule
$\pi_0$~\cite{black2026pi0visionlanguageactionflowmodel} & 15 & 55 & 55 \\
GR00T~\cite{nvidia2025gr00tn1openfoundation} & 47.5 & 20 & 15 \\
VLA-Feedback($\pi_0$) & 42.5 & \textbf{75} & 55 \\
VLA-Feedback(GR00T) & \textbf{80} & \textbf{75} & \textbf{100} \\
\bottomrule
\end{tabular}
\vspace{-10pt}
\end{wraptable}

\paragraph{VLA-Feedback Applicability.}
To evaluate applicability beyond GR00T, we also integrated VLA-Feedback with the $\pi_0$ planner. Its Transformer-based flow-matching action generator provides a compatible interface for applying feedback at the final denoising step. While Table~\ref{tab:sim_results} presents the main evaluation using GR00T as our backbone, Table~\ref{tab:pi0} evaluates this extension to $\pi_0$. Under the same evaluation protocol, VLA-Feedback increased average success across the three dynamic tasks from 27.5\% to 85.0\% with GR00T and from 41.7\% to 57.5\% with $\pi_0$, supporting applicability across both evaluated backbones.

\section{Limitations} \label{sec:limitation}

VLA-Feedback improves responsiveness within an action chunk, but it remains a local refinement method and still depends on the slow planner producing a reasonable near-final action. When the initial chunk is far from a valid solution, or when the target changes abruptly near contact, one-step feedback denoising may not have enough workspace to recover. On hardware, occlusion and actuation delay can reduce correction accuracy, while observation noise may corrupt the small feedback update. Future extensions could combine adaptive re-planning and predictive world models to refresh stale plans and compensate for noisy or delayed observations during fast feedback updates.


\section{Conclusion}
We presented VLA-Feedback, a two-timescale architecture that makes action-chunking VLAs more responsive by using the final denoising step as a fast feedback interface. Instead of rerunning the full VLM-DiT planner or learning a separate reactive policy, VLA-Feedback refines the planner's near-final action using the latest observation at control frequency. Across static LIBERO tasks, dynamic simulation tasks, and real-world manipulation experiments, VLA-Feedback largely preserved the diffusion planner's static performance while improving robustness to moving targets and execution-time changes. This feedback-denoising interface reduced estimated reaction latency and could help VLAs adapt to unpredictable environments, hardware uncertainties, and potential failures in real world deployments. More broadly, this feedback-denoising scheme provides a general mechanism for introducing high-frequency feedback into diffusion-based robot policies beyond VLAs.


\acknowledgments{This work was supported in part by the U.S. Department of the Air Force through the AFWERX Small Business Innovation Research (SBIR) Phase II program under an award to Roboligent, Inc. We would also like to thank Sandeep Chinchali for providing the robot infrastructure.

}




\bibliography{refs}

@article{shukor2025smolvla,
  title={Smolvla: A vision-language-action model for affordable and efficient robotics},
  author={Shukor, Mustafa and Aubakirova, Dana and Capuano, Francesco and Kooijmans, Pepijn and Palma, Steven and Zouitine, Adil and Aractingi, Michel and Pascal, Caroline and Russi, Martino and Marafioti, Andres and others},
  journal={arXiv preprint arXiv:2506.01844},
  year={2025}
}

@ARTICLE{10900471,
  author={Wen, Junjie and Zhu, Yichen and Li, Jinming and Zhu, Minjie and Tang, Zhibin and Wu, Kun and Xu, Zhiyuan and Liu, Ning and Cheng, Ran and Shen, Chaomin and Peng, Yaxin and Feng, Feifei and Tang, Jian},
  journal={IEEE Robotics and Automation Letters}, 
  title={{TinyVLA}: Toward Fast, Data-Efficient Vision-Language-Action Models for Robotic Manipulation}, 
  year={2025},
  volume={10},
  number={4},
  pages={3988--3995},
  doi={10.1109/LRA.2025.3544909}}

@misc{budzianowski2025edgevlaefficientvisionlanguageactionmodels,
      title={EdgeVLA: Efficient Vision-Language-Action Models}, 
      author={Paweł Budzianowski and Wesley Maa and Matthew Freed and Jingxiang Mo and Winston Hsiao and Aaron Xie and Tomasz Młoduchowski and Viraj Tipnis and Benjamin Bolte},
      year={2025},
      eprint={2507.14049},
      archivePrefix={arXiv},
      primaryClass={cs.RO},
      url={https://arxiv.org/abs/2507.14049}, 
}

@misc{dong2025vitavlaefficientlyteachingvisionlanguage,
      title={VITA-VLA: Efficiently Teaching Vision-Language Models to Act via Action Expert Distillation}, 
      author={Shaoqi Dong and Chaoyou Fu and Haihan Gao and Yi-Fan Zhang and Chi Yan and Chu Wu and Xiaoyu Liu and Yunhang Shen and Jing Huo and Deqiang Jiang and Haoyu Cao and Yang Gao and Xing Sun and Ran He and Caifeng Shan},
      year={2025},
      eprint={2510.09607},
      archivePrefix={arXiv},
      primaryClass={cs.CV},
      url={https://arxiv.org/abs/2510.09607}, 
}

@misc{yu2026surveyefficientvisionlanguageactionmodels,
      title={A Survey on Efficient Vision-Language-Action Models}, 
      author={Zhaoshu Yu and Bo Wang and Pengpeng Zeng and Haonan Zhang and Ji Zhang and Zheng Wang and Lianli Gao and Jingkuan Song and Nicu Sebe and Heng Tao Shen},
      year={2026},
      eprint={2510.24795},
      archivePrefix={arXiv},
      primaryClass={cs.CV},
      url={https://arxiv.org/abs/2510.24795}, 
}

@misc{guan2025efficientvisionlanguageactionmodelsembodied,
      title={Efficient Vision-Language-Action Models for Embodied Manipulation: A Systematic Survey}, 
      author={Weifan Guan and Qinghao Hu and Aosheng Li and Jian Cheng},
      year={2025},
      eprint={2510.17111},
      archivePrefix={arXiv},
      primaryClass={cs.RO},
      url={https://arxiv.org/abs/2510.17111}, 
}

@inproceedings{NEURIPS2025_3a2ef31a,
 author = {Yang, Yantai and Wang, Yuhao and Wen, Zichen and Zhongwei, Luo and Zou, Chang and Zhang, Zhipeng and Wen, Chuan and Zhang, Linfeng},
 booktitle = {Advances in Neural Information Processing Systems},
 doi = {10.52202/085713-1365},
 editor = {D. Belgrave and C. Zhang and H. Lin and R. Pascanu and P. Koniusz and M. Ghassemi and N. Chen},
 pages = {40891--40914},
 publisher = {Curran Associates, Inc.},
 title = {{EfficientVLA}: Training-Free Acceleration and Compression for Vision-Language-Action Models},
 url = {https://proceedings.neurips.cc/paper_files/paper/2025/file/3a2ef31a1e45908901adc0ca853a8faf-Paper-Conference.pdf},
 volume = {38, Main Conference},
 year = {2025}
}

@misc{tan2025thinktwiceactonce,
      title={Think Twice, Act Once: Token-Aware Compression and Action Reuse for Efficient Inference in Vision-Language-Action Models}, 
      author={Xudong Tan and Yaoxin Yang and Peng Ye and Jialin Zheng and Bizhe Bai and Xinyi Wang and Jia Hao and Tao Chen},
      year={2025},
      eprint={2505.21200},
      archivePrefix={arXiv},
      primaryClass={cs.CV},
      url={https://arxiv.org/abs/2505.21200}, 
}

@inproceedings{
wang2026specprunevla,
title={{SpecPrune-VLA}: Accelerating Vision-Language-Action Models via Action-Aware Self-Speculative Pruning},
author={Hanzhen Wang and Jiaming Xu and Yushun Xiang and Jiayi Pan and Yongkang Zhou and Yong-Lu Li and Guohao Dai},
booktitle={Forty-third International Conference on Machine Learning},
year={2026},
url={https://openreview.net/forum?id=MjE62ZnpRH}
}

@inproceedings{pei2026action,
  title={Action-aware dynamic pruning for efficient vision-language-action manipulation},
  author={Pei, Xiaohuan and Chen, Yuxing and Xu, Siyu and Wang, Yunke and Shi, Yuheng and Xu, Chang},
  booktitle={International Conference on Learning Representations},
  volume={2026},
  pages={10832--10851},
  year={2026}
}

@INPROCEEDINGS{11247519,
  author={Song, Wenxuan and Chen, Jiayi and Ding, Pengxiang and Zhao, Han and Zhao, Wei and Zhong, Zhide and Ge, Zongyuan and Li, Zhijun and Wang, Donglin and Wang, Lujia and Ma, Jun and Li, Haoang},
  booktitle={2025 IEEE/RSJ International Conference on Intelligent Robots and Systems (IROS)}, 
  title={{PD-VLA}: Accelerating Vision-Language-Action Model Integrated with Action Chunking via Parallel Decoding}, 
  year={2025},
  volume={},
  number={},
  pages={13162--13169},
  doi={10.1109/IROS60139.2025.11247519}}

@INPROCEEDINGS{PertschK-RSS-25, 
    AUTHOR    = {Karl Pertsch AND Kyle Stachowicz AND Brian Ichter AND Danny Driess AND Suraj Nair AND Quan Vuong AND Oier Mees AND Chelsea Finn AND Sergey Levine}, 
    TITLE     = {{FAST: Efficient Action Tokenization for Vision-Language-Action Models}}, 
    BOOKTITLE = {Proceedings of Robotics: Science and Systems}, 
    YEAR      = {2025}, 
    ADDRESS   = {Los Angeles, CA, USA}, 
    MONTH     = {June}, 
    DOI       = {10.15607/RSS.2025.XXI.012} 
}

@inproceedings{NEURIPS2025_300ccb21,
 author = {Black, Kevin and Galliker, Manuel and Levine, Sergey},
 booktitle = {Advances in Neural Information Processing Systems},
 doi = {10.52202/085713-1122},
 editor = {D. Belgrave and C. Zhang and H. Lin and R. Pascanu and P. Koniusz and M. Ghassemi and N. Chen},
 pages = {33383--33407},
 publisher = {Curran Associates, Inc.},
 title = {Real-Time Execution of Action Chunking Flow Policies},
 url = {https://proceedings.neurips.cc/paper_files/paper/2025/file/300ccb2187dedd4edcc07f7e76d8e553-Paper-Conference.pdf},
 volume = {38, Main Conference},
 year = {2025}
}

@misc{xie2026dynamicvlavisionlanguageactionmodeldynamic,
      title={DynamicVLA: A Vision-Language-Action Model for Dynamic Object Manipulation}, 
      author={Haozhe Xie and Beichen Wen and Jiarui Zheng and Zhaoxi Chen and Fangzhou Hong and Haiwen Diao and Ziwei Liu},
      year={2026},
      eprint={2601.22153},
      archivePrefix={arXiv},
      primaryClass={cs.RO},
      url={https://arxiv.org/abs/2601.22153}, 
}

@misc{jiang2026asyncvlaasynchronousflowmatching,
      title={AsyncVLA: Asynchronous Flow Matching for Vision-Language-Action Models}, 
      author={Yuhua Jiang and Shuang Cheng and Yan Ding and Feifei Gao and Biqing Qi},
      year={2026},
      eprint={2511.14148},
      archivePrefix={arXiv},
      primaryClass={cs.RO},
      url={https://arxiv.org/abs/2511.14148}, 
}

@misc{wang2026discretertcdiscretediffusionpolicies,
      title={DiscreteRTC: Discrete Diffusion Policies are Natural Asynchronous Executors}, 
      author={Pengcheng Wang and Kaiwen Hong and Chensheng Peng and Katherine Driggs-Campbell and Masayoshi Tomizuka and Chenfeng Xu and Chen Tang},
      year={2026},
      eprint={2604.25050},
      archivePrefix={arXiv},
      primaryClass={cs.RO},
      url={https://arxiv.org/abs/2604.25050}, 
}

@InProceedings{pmlr-v162-janner22a,
  title = 	 {Planning with Diffusion for Flexible Behavior Synthesis},
  author =       {Janner, Michael and Du, Yilun and Tenenbaum, Joshua and Levine, Sergey},
  booktitle = 	 {Proceedings of the 39th International Conference on Machine Learning},
  pages = 	 {9902--9915},
  year = 	 {2022},
  editor = 	 {Chaudhuri, Kamalika and Jegelka, Stefanie and Song, Le and Szepesvari, Csaba and Niu, Gang and Sabato, Sivan},
  volume = 	 {162},
  series = 	 {Proceedings of Machine Learning Research},
  month = 	 {17--23 Jul},
  publisher =    {PMLR},
  url = 	 {https://proceedings.mlr.press/v162/janner22a.html}
}

@article{chi2025diffusion,
  title={Diffusion policy: Visuomotor policy learning via action diffusion},
  author={Chi, Cheng and Xu, Zhenjia and Feng, Siyuan and Cousineau, Eric and Du, Yilun and Burchfiel, Benjamin and Tedrake, Russ and Song, Shuran},
  journal={The International Journal of Robotics Research},
  volume={44},
  number={10-11},
  pages={1684--1704},
  year={2025},
  publisher={Sage Publications Sage UK: London, England}
}

@INPROCEEDINGS{10342382,
  author={Carvalho, João and Le, An T. and Baierl, Mark and Koert, Dorothea and Peters, Jan},
  booktitle={2023 IEEE/RSJ International Conference on Intelligent Robots and Systems (IROS)}, 
  title={Motion Planning Diffusion: Learning and Planning of Robot Motions with Diffusion Models}, 
  year={2023},
  volume={},
  number={},
  pages={1916--1923},
  doi={10.1109/IROS55552.2023.10342382}}

@InProceedings{pmlr-v270-yang25a,
  title = 	 {{EquiBot}: SIM(3)-Equivariant Diffusion Policy for Generalizable and Data Efficient Learning},
  author =       {Yang, Jingyun and Cao, Ziang and Deng, Congyue and Antonova, Rika and Song, Shuran and Bohg, Jeannette},
  booktitle = 	 {Proceedings of The 8th Conference on Robot Learning},
  pages = 	 {1048--1068},
  year = 	 {2025},
  editor = 	 {Agrawal, Pulkit and Kroemer, Oliver and Burgard, Wolfram},
  volume = 	 {270},
  series = 	 {Proceedings of Machine Learning Research},
  month = 	 {06--09 Nov},
  publisher =    {PMLR},
  url = 	 {https://proceedings.mlr.press/v270/yang25a.html}
}

@inproceedings{
song2021denoising,
title={Denoising Diffusion Implicit Models},
author={Jiaming Song and Chenlin Meng and Stefano Ermon},
booktitle={International Conference on Learning Representations},
year={2021},
url={https://openreview.net/forum?id=St1giarCHLP}
}

@inproceedings{
salimans2022progressive,
title={Progressive Distillation for Fast Sampling of Diffusion Models},
author={Tim Salimans and Jonathan Ho},
booktitle={International Conference on Learning Representations},
year={2022},
url={https://openreview.net/forum?id=TIdIXIpzhoI}
}

@InProceedings{pmlr-v202-song23a,
  title = 	 {Consistency Models},
  author =       {Song, Yang and Dhariwal, Prafulla and Chen, Mark and Sutskever, Ilya},
  booktitle = 	 {Proceedings of the 40th International Conference on Machine Learning},
  pages = 	 {32211--32252},
  year = 	 {2023},
  editor = 	 {Krause, Andreas and Brunskill, Emma and Cho, Kyunghyun and Engelhardt, Barbara and Sabato, Sivan and Scarlett, Jonathan},
  volume = 	 {202},
  series = 	 {Proceedings of Machine Learning Research},
  month = 	 {23--29 Jul},
  publisher =    {PMLR},
  url = 	 {https://proceedings.mlr.press/v202/song23a.html}
}

@inproceedings{
lipman2023flow,
title={Flow Matching for Generative Modeling},
author={Yaron Lipman and Ricky T. Q. Chen and Heli Ben-Hamu and Maximilian Nickel and Matthew Le},
booktitle={The Eleventh International Conference on Learning Representations },
year={2023},
url={https://openreview.net/forum?id=PqvMRDCJT9t}
}

@INPROCEEDINGS{Prasad-RSS-24, 
    AUTHOR    = {Aaditya Prasad AND Kevin Lin AND Jimmy Wu AND Linqi Zhou AND Jeannette Bohg}, 
    TITLE     = {{Consistency Policy: Accelerated Visuomotor Policies via Consistency Distillation}}, 
    BOOKTITLE = {Proceedings of Robotics: Science and Systems}, 
    YEAR      = {2024}, 
    ADDRESS   = {Delft, Netherlands}, 
    MONTH     = {July}, 
    DOI       = {10.15607/RSS.2024.XX.071} 
}

@InProceedings{pmlr-v267-wang25ba,
  title = 	 {One-Step Diffusion Policy: Fast Visuomotor Policies via Diffusion Distillation},
  author =       {Wang, Zhendong and Li, Max and Mandlekar, Ajay and Xu, Zhenjia and Fan, Jiaojiao and Narang, Yashraj and Fan, Linxi and Zhu, Yuke and Balaji, Yogesh and Zhou, Mingyuan and Liu, Ming-Yu and Zeng, Yu},
  booktitle = 	 {Proceedings of the 42nd International Conference on Machine Learning},
  pages = 	 {63399--63416},
  year = 	 {2025},
  editor = 	 {Singh, Aarti and Fazel, Maryam and Hsu, Daniel and Lacoste-Julien, Simon and Berkenkamp, Felix and Maharaj, Tegan and Wagstaff, Kiri and Zhu, Jerry},
  volume = 	 {267},
  series = 	 {Proceedings of Machine Learning Research},
  month = 	 {13--19 Jul},
  publisher =    {PMLR},
  url = 	 {https://proceedings.mlr.press/v267/wang25ba.html}
}

@misc{black2026pi0visionlanguageactionflowmodel,
      title={$\pi_0$: A Vision-Language-Action Flow Model for General Robot Control}, 
      author={Kevin Black and Noah Brown and Danny Driess and Adnan Esmail and Michael Equi and Chelsea Finn and Niccolo Fusai and Lachy Groom and Karol Hausman and Brian Ichter and Szymon Jakubczak and Tim Jones and Liyiming Ke and Sergey Levine and Adrian Li-Bell and Mohith Mothukuri and Suraj Nair and Karl Pertsch and Lucy Xiaoyang Shi and James Tanner and Quan Vuong and Anna Walling and Haohuan Wang and Ury Zhilinsky},
      year={2026},
      eprint={2410.24164},
      archivePrefix={arXiv},
      primaryClass={cs.LG},
      url={https://arxiv.org/abs/2410.24164}, 
}

@inproceedings{ICLR2025_49f80e4d,
 author = {Liu, Songming and Wu, Lingxuan and Li, Bangguo and Tan, Hengkai and Chen, Huayu and Wang, Zhengyi and Xu, Ke and Su, Hang and Zhu, Jun},
 booktitle = {International Conference on Learning Representations},
 editor = {Y. Yue and A. Garg and N. Peng and F. Sha and R. Yu},
 pages = {29982--30009},
 title = {{RDT-1B}: a Diffusion Foundation Model for Bimanual Manipulation},
 url = {https://proceedings.iclr.cc/paper_files/paper/2025/file/49f80e4d2471ad4f2edf4f5f1ab62339-Paper-Conference.pdf},
 volume = {2025},
 year = {2025}
}

@article{nvidia2025gr00tn1openfoundation,
  title={Gr00t n1: An open foundation model for generalist humanoid robots},
  author={Bjorck, Johan and Casta{\~n}eda, Fernando and Cherniadev, Nikita and Da, Xingye and Ding, Runyu and Fan, Linxi and Fang, Yu and Fox, Dieter and Hu, Fengyuan and Huang, Spencer and others},
  journal={arXiv preprint arXiv:2503.14734},
  year={2025}
}

@InProceedings{pmlr-v305-wen25b,
  title = 	 {{DexVLA}: Vision-Language Model with Plug-In Diffusion Expert for General Robot Control},
  author =       {Wen, Junjie and Zhu, Yichen and Li, Jinming and Tang, Zhibin and Shen, Chaomin and Feng, Feifei},
  booktitle = 	 {Proceedings of The 9th Conference on Robot Learning},
  pages = 	 {3094--3114},
  year = 	 {2025},
  editor = 	 {Lim, Joseph and Song, Shuran and Park, Hae-Won},
  volume = 	 {305},
  series = 	 {Proceedings of Machine Learning Research},
  month = 	 {27--30 Sep},
  publisher =    {PMLR},
  url = 	 {https://proceedings.mlr.press/v305/wen25b.html}
}

@InProceedings{Hou_2025_ICCV,
    author    = {Hou, Zhi and Zhang, Tianyi and Xiong, Yuwen and Duan, Haonan and Pu, Hengjun and Tong, Ronglei and Zhao, Chengyang and Zhu, Xizhou and Qiao, Yu and Dai, Jifeng and Chen, Yuntao},
    title     = {{Dita}: Scaling Diffusion Transformer for Generalist Vision-Language-Action Policy},
    booktitle = {Proceedings of the IEEE/CVF International Conference on Computer Vision (ICCV)},
    month     = {October},
    year      = {2025},
    pages     = {7686--7697}
}

@article{ho2022classifier,
  title={Classifier-free diffusion guidance},
  author={Ho, Jonathan and Salimans, Tim},
  journal={arXiv preprint arXiv:2207.12598},
  year={2022}
}

@InProceedings{Zhang_2023_ICCV,
    author    = {Zhang, Lvmin and Rao, Anyi and Agrawala, Maneesh},
    title     = {Adding Conditional Control to Text-to-Image Diffusion Models},
    booktitle = {Proceedings of the IEEE/CVF International Conference on Computer Vision (ICCV)},
    month     = {October},
    year      = {2023},
    pages     = {3836--3847}
}

@InProceedings{pmlr-v305-li25c,
  title = 	 {{ControlVLA}: Few-shot Object-centric Adaptation for Pre-trained Vision-Language-Action Models},
  author =       {Li, Puhao and Wu, Yingying and Xi, Ziheng and Li, Wanlin and Huang, Yuzhe and Zhang, Zhiyuan and Chen, Yinghan and Wang, Jianan and Zhu, Song-Chun and Liu, Tengyu and Huang, Siyuan},
  booktitle = 	 {Proceedings of The 9th Conference on Robot Learning},
  pages = 	 {1898--1913},
  year = 	 {2025},
  editor = 	 {Lim, Joseph and Song, Shuran and Park, Hae-Won},
  volume = 	 {305},
  series = 	 {Proceedings of Machine Learning Research},
  month = 	 {27--30 Sep},
  publisher =    {PMLR},
  url = 	 {https://proceedings.mlr.press/v305/li25c.html}
}

@INPROCEEDINGS{11272538,
  author={Lu, Yuang and Wang, Song and Han, Xiao and Zhang, Xuri and Wu, Yucong and He, Zhicheng},
  booktitle={2025 IEEE 37th International Conference on Tools with Artificial Intelligence (ICTAI)}, 
  title={Enhancing Diffusion Policy with Classifier-Free Guidance for Temporal Robotic Tasks}, 
  year={2025},
  volume={},
  number={},
  pages={1023--1029},
  doi={10.1109/ICTAI66417.2025.00150}}

@ARTICLE{10912754,
  author={Wang, Dexin and Liu, Chunsheng and Chang, Faliang and Xu, Yichen},
  journal={IEEE Transactions on Robotics}, 
  title={Hierarchical Diffusion Policy: Manipulation Trajectory Generation via Contact Guidance}, 
  year={2025},
  volume={41},
  number={},
  pages={2086--2104},
  doi={10.1109/TRO.2025.3547272}}

@INPROCEEDINGS{11127231,
  author={Li, Hang and Feng, Qian and Zheng, Zhi and Feng, Jianxiang and Chen, Zhaopeng and Knoll, Alois},
  booktitle={2025 IEEE International Conference on Robotics and Automation (ICRA)}, 
  title={Language-Guided Object-Centric Diffusion Policy for Generalizable and Collision-Aware Manipulation}, 
  year={2025},
  volume={},
  number={},
  pages={12834--12841},
  doi={10.1109/ICRA55743.2025.11127231}}

@INPROCEEDINGS{Brohan-RSS-23, 
    AUTHOR    = {Anthony Brohan AND Noah Brown AND Justice Carbajal AND Yevgen Chebotar AND Joseph Dabis AND Chelsea Finn AND Keerthana Gopalakrishnan AND Karol Hausman AND Alexander Herzog AND Jasmine Hsu AND Julian Ibarz AND Brian Ichter AND Alex Irpan AND Tomas Jackson AND Sally Jesmonth AND Nikhil Joshi AND Ryan Julian AND Dmitry Kalashnikov AND Yuheng Kuang AND Isabel Leal AND Kuang-Huei Lee AND Sergey Levine AND Yao Lu AND Utsav Malla AND Deeksha Manjunath AND Igor Mordatch AND Ofir Nachum AND Carolina Parada AND Jodilyn  Peralta AND Emily Perez AND Karl Pertsch AND Jornell  Quiambao AND Kanishka Rao AND Michael S Ryoo AND Grecia  Salazar AND Pannag R Sanketi AND Kevin  Sayed AND Jaspiar  Singh AND Sumedh  Sontakke AND Austin  Stone AND Clayton  Tan AND Huong  Tran AND Vincent Vanhoucke AND Steve  Vega AND Quan H Vuong AND Fei Xia AND Ted Xiao AND Peng Xu AND Sichun Xu AND Tianhe Yu AND Brianna  Zitkovich}, 
    TITLE     = {{RT-1: Robotics Transformer for Real-World Control at Scale}}, 
    BOOKTITLE = {Proceedings of Robotics: Science and Systems}, 
    YEAR      = {2023}, 
    ADDRESS   = {Daegu, Republic of Korea}, 
    MONTH     = {July}, 
    DOI       = {10.15607/RSS.2023.XIX.025} 
}

@InProceedings{pmlr-v229-zitkovich23a,
  title = 	 {{RT-2}: Vision-Language-Action Models Transfer Web Knowledge to Robotic Control},
  author =       {Zitkovich, Brianna and Yu, Tianhe and Xu, Sichun and Xu, Peng and Xiao, Ted and Xia, Fei and Wu, Jialin and Wohlhart, Paul and Welker, Stefan and Wahid, Ayzaan and Vuong, Quan and Vanhoucke, Vincent and Tran, Huong and Soricut, Radu and Singh, Anikait and Singh, Jaspiar and Sermanet, Pierre and Sanketi, Pannag R. and Salazar, Grecia and Ryoo, Michael S. and Reymann, Krista and Rao, Kanishka and Pertsch, Karl and Mordatch, Igor and Michalewski, Henryk and Lu, Yao and Levine, Sergey and Lee, Lisa and Lee, Tsang-Wei Edward and Leal, Isabel and Kuang, Yuheng and Kalashnikov, Dmitry and Julian, Ryan and Joshi, Nikhil J. and Irpan, Alex and Ichter, Brian and Hsu, Jasmine and Herzog, Alexander and Hausman, Karol and Gopalakrishnan, Keerthana and Fu, Chuyuan and Florence, Pete and Finn, Chelsea and Dubey, Kumar Avinava and Driess, Danny and Ding, Tianli and Choromanski, Krzysztof Marcin and Chen, Xi and Chebotar, Yevgen and Carbajal, Justice and Brown, Noah and Brohan, Anthony and Arenas, Montserrat Gonzalez and Han, Kehang},
  booktitle = 	 {Proceedings of The 7th Conference on Robot Learning},
  pages = 	 {2165--2183},
  year = 	 {2023},
  editor = 	 {Tan, Jie and Toussaint, Marc and Darvish, Kourosh},
  volume = 	 {229},
  series = 	 {Proceedings of Machine Learning Research},
  month = 	 {06--09 Nov},
  publisher =    {PMLR},
  url = 	 {https://proceedings.mlr.press/v229/zitkovich23a.html}
}

@InProceedings{pmlr-v270-kim25c,
  title = 	 {{OpenVLA}: An Open-Source Vision-Language-Action Model},
  author =       {Kim, Moo Jin and Pertsch, Karl and Karamcheti, Siddharth and Xiao, Ted and Balakrishna, Ashwin and Nair, Suraj and Rafailov, Rafael and Foster, Ethan P and Sanketi, Pannag R and Vuong, Quan and Kollar, Thomas and Burchfiel, Benjamin and Tedrake, Russ and Sadigh, Dorsa and Levine, Sergey and Liang, Percy and Finn, Chelsea},
  booktitle = 	 {Proceedings of The 8th Conference on Robot Learning},
  pages = 	 {2679--2713},
  year = 	 {2025},
  editor = 	 {Agrawal, Pulkit and Kroemer, Oliver and Burgard, Wolfram},
  volume = 	 {270},
  series = 	 {Proceedings of Machine Learning Research},
  month = 	 {06--09 Nov},
  publisher =    {PMLR},
  url = 	 {https://proceedings.mlr.press/v270/kim25c.html}
}

@INPROCEEDINGS{Ghosh-RSS-24, 
    AUTHOR    = {Dibya Ghosh AND Homer Rich Walke AND Karl Pertsch AND Kevin Black AND Oier Mees AND Sudeep Dasari AND Joey Hejna AND Tobias Kreiman AND Charles Xu AND Jianlan Luo AND You Liang Tan AND Lawrence Yunliang Chen AND Quan Vuong AND Ted Xiao AND Pannag R Sanketi AND Dorsa Sadigh AND Chelsea Finn AND Sergey Levine}, 
    TITLE     = {{Octo: An Open-Source Generalist Robot Policy}}, 
    BOOKTITLE = {Proceedings of Robotics: Science and Systems}, 
    YEAR      = {2024}, 
    ADDRESS   = {Delft, Netherlands}, 
    MONTH     = {July}, 
    DOI       = {10.15607/RSS.2024.XX.090} 
}

@article{song2025hume,
  title={Hume: Introducing system-2 thinking in visual-language-action model},
  author={Song, Haoming and Qu, Delin and Yao, Yuanqi and Chen, Qizhi and Lv, Qi and Tang, Yiwen and Shi, Modi and Ren, Guanghui and Yao, Maoqing and Zhao, Bin and others},
  journal={arXiv preprint arXiv:2505.21432},
  year={2025}
}

@InProceedings{pmlr-v270-zhang25b,
  title = 	 {{HiRT}: Enhancing Robotic Control with Hierarchical Robot Transformers},
  author =       {Zhang, Jianke and Guo, Yanjiang and Chen, Xiaoyu and Wang, Yen-Jen and Hu, Yucheng and Shi, Chengming and Chen, Jianyu},
  booktitle = 	 {Proceedings of The 8th Conference on Robot Learning},
  pages = 	 {933--946},
  year = 	 {2025},
  editor = 	 {Agrawal, Pulkit and Kroemer, Oliver and Burgard, Wolfram},
  volume = 	 {270},
  series = 	 {Proceedings of Machine Learning Research},
  month = 	 {06--09 Nov},
  publisher =    {PMLR},
  url = 	 {https://proceedings.mlr.press/v270/zhang25b.html}
}

@article{han2024dual,
  title={A dual process vla: Efficient robotic manipulation leveraging vlm},
  author={Han, ByungOk and Kim, Jaehong and Jang, Jinhyeok},
  journal={arXiv preprint arXiv:2410.15549},
  year={2024}
}

@article{chen2025fast,
  title={Fast-in-slow: A dual-system foundation model unifying fast manipulation within slow reasoning},
  author={Chen, Hao and Liu, Jiaming and Gu, Chenyang and Liu, Zhuoyang and Zhang, Renrui and Li, Xiaoqi and He, Xiao and Guo, Yandong and Fu, Chi-Wing and Zhang, Shanghang and others},
  journal={arXiv preprint arXiv:2506.01953},
  year={2025}
}

@article{bu2024towards,
  title={Towards synergistic, generalized, and efficient dual-system for robotic manipulation},
  author={Bu, Qingwen and Li, Hongyang and Chen, Li and Cai, Jisong and Zeng, Jia and Cui, Heming and Yao, Maoqing and Qiao, Yu},
  journal={arXiv preprint arXiv:2410.08001},
  year={2024}
}

@inproceedings{
xiong2026hypervla,
title={{HyperVLA}: Efficient Inference in Vision-Language-Action Models via Hypernetworks},
author={Zheng Xiong and Kang Li and Zilin Wang and Matthew Thomas Jackson and Jakob Nicolaus Foerster and Shimon Whiteson},
booktitle={The Fourteenth International Conference on Learning Representations},
year={2026},
url={https://openreview.net/forum?id=bsXkBTZjgY}
}

@article{sendai2025leave,
  title={Leave no observation behind: Real-time correction for vla action chunks},
  author={Sendai, Kohei and Alvarez, Maxime and Matsushima, Tatsuya and Matsuo, Yutaka and Iwasawa, Yusuke},
  journal={arXiv preprint arXiv:2509.23224},
  year={2025}
}

@inproceedings{
yuan2025policy,
title={Policy Decorator: Model-Agnostic Online Refinement for Large Policy Model},
author={Xiu Yuan and Tongzhou Mu and Stone Tao and Yunhao Fang and Mengke Zhang and Hao Su},
booktitle={The Thirteenth International Conference on Learning Representations},
year={2025},
url={https://openreview.net/forum?id=e5jGTEiJMT}
}

@article{jiang2026fast,
  title={How fast can i run my vla? demystifying vla inference performance with vla-perf},
  author={Jiang, Wenqi and Clemons, Jason and Sankaralingam, Karu and Kozyrakis, Christos},
  journal={arXiv preprint arXiv:2602.18397},
  year={2026}
}

@misc{agouzoul2026understandingasynchronousinferencemethods,
      title={Understanding Asynchronous Inference Methods for Vision-Language-Action Models}, 
      author={Ayoub Agouzoul},
      year={2026},
      eprint={2605.08168},
      archivePrefix={arXiv},
      primaryClass={cs.RO},
      url={https://arxiv.org/abs/2605.08168}, 
}

@misc{liang2026adaptiveactionchunkinginferencetime,
      title={Adaptive Action Chunking at Inference-time for Vision-Language-Action Models}, 
      author={Yuanchang Liang and Xiaobo Wang and Kai Wang and Shuo Wang and Xiaojiang Peng and Haoyu Chen and David Kim Huat Chua and Prahlad Vadakkepat},
      year={2026},
      eprint={2604.04161},
      archivePrefix={arXiv},
      primaryClass={cs.RO},
      url={https://arxiv.org/abs/2604.04161}, 
}

@misc{lu2026fasterrethinkingrealtimeflow,
      title={FASTER: Rethinking Real-Time Flow VLAs}, 
      author={Yuxiang Lu and Zhe Liu and Xianzhe Fan and Zhenya Yang and Jinghua Hou and Junyi Li and Kaixin Ding and Hengshuang Zhao},
      year={2026},
      eprint={2603.19199},
      archivePrefix={arXiv},
      primaryClass={cs.RO},
      url={https://arxiv.org/abs/2603.19199}, 
}

@misc{tang2025vlashrealtimevlasfuturestateaware,
      title={VLASH: Real-Time VLAs via Future-State-Aware Asynchronous Inference}, 
      author={Jiaming Tang and Yufei Sun and Yilong Zhao and Shang Yang and Yujun Lin and Zhuoyang Zhang and James Hou and Yao Lu and Zhijian Liu and Song Han},
      year={2025},
      eprint={2512.01031},
      archivePrefix={arXiv},
      primaryClass={cs.RO},
      url={https://arxiv.org/abs/2512.01031}, 
}

@misc{li2025eagle2buildingposttraining,
      title={Eagle 2: Building Post-Training Data Strategies from Scratch for Frontier Vision-Language Models}, 
      author={Zhiqi Li and Guo Chen and Shilong Liu and Shihao Wang and Vibashan VS and Yishen Ji and Shiyi Lan and Hao Zhang and Yilin Zhao and Subhashree Radhakrishnan and Nadine Chang and Karan Sapra and Amala Sanjay Deshmukh and Tuomas Rintamaki and Matthieu Le and Ilia Karmanov and Lukas Voegtle and Philipp Fischer and De-An Huang and Timo Roman and Tong Lu and Jose M. Alvarez and Bryan Catanzaro and Jan Kautz and Andrew Tao and Guilin Liu and Zhiding Yu},
      year={2025},
      eprint={2501.14818},
      archivePrefix={arXiv},
      primaryClass={cs.CV},
      url={https://arxiv.org/abs/2501.14818}, 
}

@InProceedings{Peebles_2023_ICCV,
    author    = {Peebles, William and Xie, Saining},
    title     = {Scalable Diffusion Models with Transformers},
    booktitle = {Proceedings of the IEEE/CVF International Conference on Computer Vision (ICCV)},
    month     = {October},
    year      = {2023},
    pages     = {4195--4205}
}

@inproceedings{NIPS2017_3f5ee243,
 author = {Vaswani, Ashish and Shazeer, Noam and Parmar, Niki and Uszkoreit, Jakob and Jones, Llion and Gomez, Aidan N and Kaiser, \L ukasz and Polosukhin, Illia},
 booktitle = {Advances in Neural Information Processing Systems},
 editor = {I. Guyon and U. Von Luxburg and S. Bengio and H. Wallach and R. Fergus and S. Vishwanathan and R. Garnett},
 pages = {5998--6008},
 publisher = {Curran Associates, Inc.},
 title = {Attention is All you Need},
 url = {https://proceedings.neurips.cc/paper_files/paper/2017/file/3f5ee243547dee91fbd053c1c4a845aa-Paper.pdf},
 volume = {30},
 year = {2017}
}

@article{liu2023libero,
  title={Libero: Benchmarking knowledge transfer for lifelong robot learning},
  author={Liu, Bo and Zhu, Yifeng and Gao, Chongkai and Feng, Yihao and Liu, Qiang and Zhu, Yuke and Stone, Peter},
  journal={Advances in Neural Information Processing Systems},
  volume={36},
  pages={44776--44791},
  year={2023}
}

@article{zhu2020robosuite,
  title={robosuite: A modular simulation framework and benchmark for robot learning},
  author={Zhu, Yuke and Wong, Josiah and Mandlekar, Ajay and Mart{\'\i}n-Mart{\'\i}n, Roberto and Joshi, Abhishek and Lin, Kevin and Maddukuri, Abhiram and Nasiriany, Soroush and Zhu, Yifeng},
  journal={arXiv preprint arXiv:2009.12293},
  year={2020}
}

@book{kahneman2011thinking,
  title={Thinking, fast and slow},
  author={Kahneman, Daniel},
  year={2011},
  publisher={macmillan}
}

\newpage

\appendix

\section{Appendix}

\subsection{Task Design and Evaluation Protocol}
\label{app:tasks}

\paragraph{Dynamic Simulation Tasks.}

We evaluated three dynamic Robosuite tasks. Simulation velocities switched discretely every 2~s rather than being generated from a prescribed acceleration range. \textbf{\textit{Pick up 1D Moving Robot.}} The robot grasped a toy robot moving along one axis, with velocity $v_y \in \{0, -0.04\}\,\mathrm{m/s}$. \textbf{\textit{Pick up 2D Moving Robot.}} The robot grasped a toy robot whose velocity changed along both axes, with $v_x \in \{0, \pm 0.012\}\,\mathrm{m/s}$ and $v_y \in \{0, -0.038\}\,\mathrm{m/s}$. \textbf{\textit{Drop Ball to Moving Cup.}} The robot released a ball into a cup mounted on a cart moving at $v_y = -0.05\,\mathrm{m/s}$. For the 1D and 2D moving-robot tasks, success required catching and lifting the toy robot by 10~cm within 60~s. For the ball-dropping task, success required dropping the ball into the cup.

\paragraph{Unseen Dynamic Variants.}

We evaluated four unseen dynamic variants. In \textbf{\textit{Catch Block}}, the robot caught a block of similar size to the toy robot, placed on a moving cart that used the in-distribution 1D velocity set $v_y \in \{0, -0.04\}\,\mathrm{m/s}$ but followed a different fixed motion pattern. In \textbf{\textit{1D Robot Speed}}, the target followed the same one-dimensional moving-robot task, but its velocity switched among the unseen values $v_y \in \{0, -0.03, -0.05\}\,\mathrm{m/s}$. In \textbf{\textit{Drop Ball Color}}, the cup color was changed while its speed remained $v_y = -0.05\,\mathrm{m/s}$. In \textbf{\textit{Drop Ball Speed}}, the cart moved at the unseen speed $v_y = -0.066\,\mathrm{m/s}$. Success required either lifting the target object by 10~cm or dropping the ball into the cup within 60~s, depending on the task.

\paragraph{Real-World Tasks.}

We evaluated one static task and two dynamic tasks on a Franka Emika Panda robot. In \textbf{\textit{Pick up Bread}}, success required grasping and lifting the stationary bread. In \textbf{\textit{Catch the rolling can}}, the can was given an impulse push from the table edge with an initial speed of approximately $5$--$10\,\mathrm{cm/s}$ and then rolled freely; success required grasping and lifting the can. In \textbf{\textit{Drop the lemonade into the cup}}, the cup was pulled by a string at approximately $5$--$10\,\mathrm{cm/s}$; success required dropping the lemonade into the moving cup. GR00T and VLA-Feedback used the same initial pose ranges, motion procedures, rollout budgets, and success criteria.

\paragraph{Speed Robustness.}

\begin{wraptable}{r}{0.32\textwidth}
\centering
\caption{Drop-ball success rate versus speed.}
\label{tab:drop_ball_speed}
\scriptsize
\setlength{\tabcolsep}{5pt}
\renewcommand{\arraystretch}{1.05}
\begin{tabular}{@{}lcc@{}}
\toprule
Speed & GR00T & VLA-Feedback \\
\midrule
Original & 15\% & 100\% \\
$+10\%$ & 5\% & 92.5\% \\
$+20\%$ & 5\% & 85\% \\
$+30\%$ & 0\% & 57.5\% \\
$+40\%$ & 0\% & 5\% \\
\bottomrule
\end{tabular}
\end{wraptable}

Table~\ref{tab:drop_ball_speed} evaluated robustness as the target speed increased beyond the training setting. VLA-Feedback remained substantially more robust than GR00T across moderate speed increases, although performance degraded for both methods as the target motion became more challenging. More broadly, the gains on unseen variants indicated improved execution-time robustness: changes in motion, speed, or appearance could make an open-loop chunk stale or misaligned, while current observations allowed VLA-Feedback to update the near-final action.

\subsection{Model and Training Protocol}
\label{app:baselines}

\paragraph{Feedback Module Implementation Details.}

For the default GR00T-based implementation, the feedback module uses a frozen ImageNet-pretrained ResNet-18 (\(\sim\!11.2\)M parameters; 64 visual tokens), linear action and visual projections to \(d=256\), one pre-norm Transformer decoder layer (4 heads, FFN dimension 512), a residual gate initialized to 0.3, and a \(256\!\to\!7\) velocity head. Only \(\sim\!1.19\)M parameters are trainable, with \(\sim\!12.4\)M total parameters including the frozen visual encoder, compared with the \(\sim\!3\)B frozen planner.

\paragraph{Feedback Module Training Strategy.}

We explored end-to-end training during development, but it frequently failed to converge and was substantially more expensive because gradients propagated through the full VLM-DiT. As these exploratory runs were not conducted under our final evaluation protocol, we did not report them quantitatively. We therefore froze the converged planner and trained only the feedback module, providing a stable near-final action distribution. We hypothesized that this also reduced interference between long-horizon planning and local correction objectives.

\begin{table}[t]
\centering
\caption{Baseline implementation details. Schedule denotes slow-planner calls, fast-module updates, and executed actions per chunk.}
\label{tab:baseline_setup_app}
\resizebox{\linewidth}{!}{
\begin{tabular}{lccccc}
\toprule
Method & Backbone / Head & Obs. & Horizon & Schedule & Training / Notes \\
\midrule
OpenVLA~\cite{pmlr-v270-kim25c}
& Transformer VLA action head
& Agent
& 1
& $1{:}0{:}1$
& Released finetuning; agent-view setup. \\
GR00T~\cite{nvidia2025gr00tn1openfoundation}
& VLM-DiT flow-matching head
& Agent + hand
& 16
& $1{:}0{:}16$
& Released finetuning; open-loop chunks. \\
FiS-VLA~\cite{chen2025fast}
& Slow--fast VLA pathways
& Agent + hand
& 1
& $1{:}4{:}4$
& Frozen vision backbone; LoRA language finetuning; released default horizon. \\
VLA-Feedback
& VLM-DiT + feedback denoising
& Agent + hand
& 16
& $1{:}16{:}16$
& Same planner as GR00T; feedback final denoising. \\
\bottomrule
\end{tabular}
}
\end{table}

\paragraph{Baseline Training Protocol.}

Table~\ref{tab:baseline_setup_app} summarized the baseline implementations. GR00T, FiS-VLA, and VLA-Feedback used both agent-view and hand-view observations when supported, while OpenVLA used only the agent view following its released setup. VLA-Feedback used both views in the slow planner and only the hand-view observation in the fast feedback decoder. All methods were trained until validation loss stabilized. FiS-VLA used its official action horizon of 1 following the released implementation; due to GPU memory limits, we froze its vision backbone and applied LoRA finetuning to the language backbone. For real-world experiments, GR00T and VLA-Feedback used the same 50 demonstrations per task and were evaluated under the same 20 rollout conditions.

\paragraph{$\pi_0$ Implementation.} We initialized $\pi_0$~\cite{black2026pi0visionlanguageactionflowmodel} from the released \texttt{pi0\_base} weights and finetuned it on our demonstrations for 10k steps with the vision-language backbone frozen. The baseline and feedback variant shared the same finetuned planner, observation inputs, action horizon $H=16$, and control frequency of 10~Hz. We adapted the feedback denoising module to $\pi_0$'s action representation and denoising convention, using the near-final action chunk, cached action features, and latest wrist image to perform the final denoising update. The feedback module followed the same design and training objective as the GR00T-based implementation, with the planner frozen during feedback training. Both variants used identical evaluation resets, seeds, and episode counts (Table~\ref{tab:pi0}).

\subsection{Detailed Latency Analysis}
\label{app:latency}

Table~\ref{tab:feedback_latency} reported the full per-step feedback path. On hardware, camera latency included exposure, acquisition, and delivery to Python, while IPC included serialization and ZMQ transfer. The measured end-to-end latency was 71.01\,ms, with only 1.95\,ms from feedback inference; most remaining latency came from sensing and communication. Feedback and control ran at 10\,Hz (100\,ms interval), while the slow planner was invoked once per 16-action chunk (0.625\,Hz), so the feedback path fit within one control interval without reducing the configured control rate. No intra-chunk replanning was used. In simulation, physical sensing, robot communication, and actuation were absent, so the reported 2.04\,ms reflected only the software-side feedback path.

\begin{table}[t]
\centering
\caption{Feedback-loop latency in simulation and hardware. All values are in milliseconds.}
\label{tab:feedback_latency}
\scriptsize
\setlength{\tabcolsep}{5pt}
\renewcommand{\arraystretch}{1.05}
\begin{tabular}{@{}lcc@{}}
\toprule
Component & Simulation & Hardware \\
\midrule
Camera capture/exposure & n/a & 62.00 \\
IPC & 0.03 & 4.00 \\
Preprocessing & 0.06 & 0.06 \\
Feedback inference & 1.95 & 1.95 \\
Robot communication & n/a & 2.00 \\
Actuation & n/a & 1.00 \\
\midrule
Total & 2.04 & 71.01 \\
\bottomrule
\end{tabular}
\end{table}


\end{document}